# SEMI-AUTOMATIC CONTROL OF TELEPRESENCE ROBOTS

Dmitry SUVOROV, Roman ZHUKOV, Dzmitry TSETSERUKOU, Anton EVMENENKO,
Boris GOLDSTEIN

Space Robotics Lab., Skolkovo Institute of Science and Technology
Robotics Department, Bauman Moscow State Technical University
Moscow, Russia

dmitry.suvorov@skolkovotech.ru, D.Tsetserukou@skoltech.ru, roman.zhukov@skolkovotech.ru,
anton.evmenenko.93@gmail.com, drgold4you@gmail.com

**Abstract:** Companies all over the world started to produce and sell telepresence robots in the last decade. Users of these robots have to plan their way to avoid collisions and control the speed of the robot to move it to destination point. This interface is familiar to many people who play computer games, but it causes difficulties for the older generation and disabled people. We present here a novel interface to control telepresence robots where the operator only specifies the destination point on the video frame and the robot moves there automatically and avoids collisions. Users of the robots learn how to perform tasks 7-28% faster and make 17-49% fewer mistakes. The approach can be easily applied for any commercial telepresence robot and for development of telepresence robots for disabled people.

## 1. INTRODUCTION

Telepresence robots were one of the first robots that entered the mass market. They became the next generation of teleconferencing systems. Over the past decade, the quest to develop a user-friendly interface for telepresence robots (Do et al., 2013; Tsetserukou et al., 2007) and special interfaces for disabled people (Tonin et al., 2011; Leeb et al., 2015) has motivated significant research efforts.

There are three categories of previously developed approaches to telepresence robot control. People have to use keyboard or joystick to control translational and rotational speeds of the robot when the robot has the first category interface. The drawback of this approach is that users have to plan robots' trajectories manually and to avoid collisions with dynamic objects. Researchers solved it in the second approach where the operator specifies the destination point for the robot on the map of the environment using mouse or touch screen (Kwon et al., 2010). Now the robot cannot operate in an unfamiliar place because it needs the environment map. The robot can build a map using LIDAR (Thrun et al., 2000), video cameras or their combination (Forsyth and Ponce, 2003). The third approach combines different methods of robot control for disabled people. For example, researchers have tried to combine the first approach with brain computer interface (Tonin et al., 2011).

In this paper, we demonstrate a new approach to telepresence robot control. It is based on the kinematic and optical model of robot and on a purely reactive approach to the navigation problem (Blanco et al., 2006). Therefore, the robot can operate in an unfamiliar place. Operators only specify destination point on the video frame and the robot travels there automatically. The purely reactive approach to the navigation problem was originally designed for autonomous mobile robots with precise localization system, for example, based on LIDAR, odometry and particle filtering. We demonstrate here that it can extend the user interface of telepresence robots without precise localization system.

## 2. OVERVIEW OF THE APPROACH

All telepresence robots have at least one video camera. Some of them have LIDAR and odometry system installed onboard. We shall consider only such robots here. The operator has to specify destination point $P_d^F$ on the video frame in the proposed approach. Next, the system needs to convert the image coordinates of the destination point into the relative coordinates of the robot's platform $P_d^R$. The last step is automatic travel of the robot to the destination point $P_d^R$.

The optical model of the robot's camera can be described as shown in equation 1 (Thrun et al., 2000).

$$\begin{pmatrix} x \\ y \\ \omega \end{pmatrix} = \begin{pmatrix} f_x & 0 & c_x \\ 0 & f_y & c_y \\ 0 & 0 & 1 \end{pmatrix} \cdot \begin{pmatrix} X \\ Y \\ Z \end{pmatrix} \quad (1)$$

It is necessary to expand equation 1 to take into account the distortion of the lens. We get equation 2

$$\begin{cases} x' = x \cdot (1 + k_1 \cdot r^2 + k_2 \cdot r^4 + k_3 \cdot r^6) + 2 \cdot p_1 \cdot x \cdot y + p_2 \cdot (r^2 + 2 \cdot x^2) \\ y' = y \cdot (1 + k_1 \cdot r^2 + k_2 \cdot r^4 + k_3 \cdot r^6) + p_1 \cdot (r^2 + 2 \cdot y^2) + 2 \cdot p_2 \cdot x \cdot y \end{cases} \quad (2)$$

where
$(x, y)^T$ are the coordinates of the projection of the point on the camera's matrix with ideal lens;

$(x', y')^T$ the coordinates of the projection of the point on the camera's matrix with real lens;

$(X, Y, Z)^T$ the coordinates of the point in the camera's coordinate system;

$f_x, f_y$ - focal lengths;

$c_x, c_y$ - the coordinates of the optical center of the cameras lens on the camera's matrix;

$(k_1, k_2, k_3)^T$ radial distortion coefficients;

$(p_1, p_2)^T$ tangential distortion coefficients.

$f_x, f_y, c_x, c_y, (k_1, k_2, k_3)^T, (p_1, p_2)^T$ are determined using the procedure of camera calibration.

When the user specifies $(x', y')^T$ using mouse or touch screen to move the robot there. there is an infinite number of $(X, Y, Z, 1)^T$ which can be the solution of equation (1). All these solutions lie on the same line L. This line L passes through the optical center of the camera that has the coordinates of $P_0 = (0,0,0,1)^T$ in the camera's coordinate system. Another point of the line $P_1 = (X, Y, Z, 1)$ can be calculated from (1) and (2) assuming Z = 1.

Next, we need to transform these points from the coordinate system associated with the camera to the coordinate system associated with the base of the robot to use them for calculation of the real destination point. So we need to describe the camera position using Denavit-Hartenberg parameters $q_1 \dots q_n$ and calculate transition matrix $T(q_1 \dots q_n)$ (Craig, 1955). The coordinates of $P_0$ and $P_1$ in the coordinate system associated with the robot's base are as in equations 3 and 4.

$$P_0^R(q_1 \dots q_n) = T(q_1 \dots q_n) \cdot \begin{pmatrix} 0 \\ 0 \\ 0 \\ 1 \end{pmatrix} \quad (3)$$

$$P_1^R(x, y, q_1 \dots q_n) = T(q_1 \dots q_n) * P_1(x, y) \quad (4)$$

The parametric form of the equation of line L (5):
$$P(t) = P_0^R + t * (P_1^R - P_0^R) \quad (5)$$

We need to assume that Z = 0 in equation 5 to calculate destination point $P_d^R$ in the coordinate system associated with base of the robot.

We use reactive navigation system described by J.-L. Blanco, J. Gonzlez and J.-A. Fernandez-Madrigal (2006) to move the robot to destination point $P_d^R$. We use odometry as a localization system.

## 3. IMPLEMENTATION

The approach was implemented for the telepresence robot Webot (Fig. 1) equipped with the RPLidar laser scanner. Its kinematic chain configuration described using Denavit-Hartenberg parameters is shown in Figure 2.

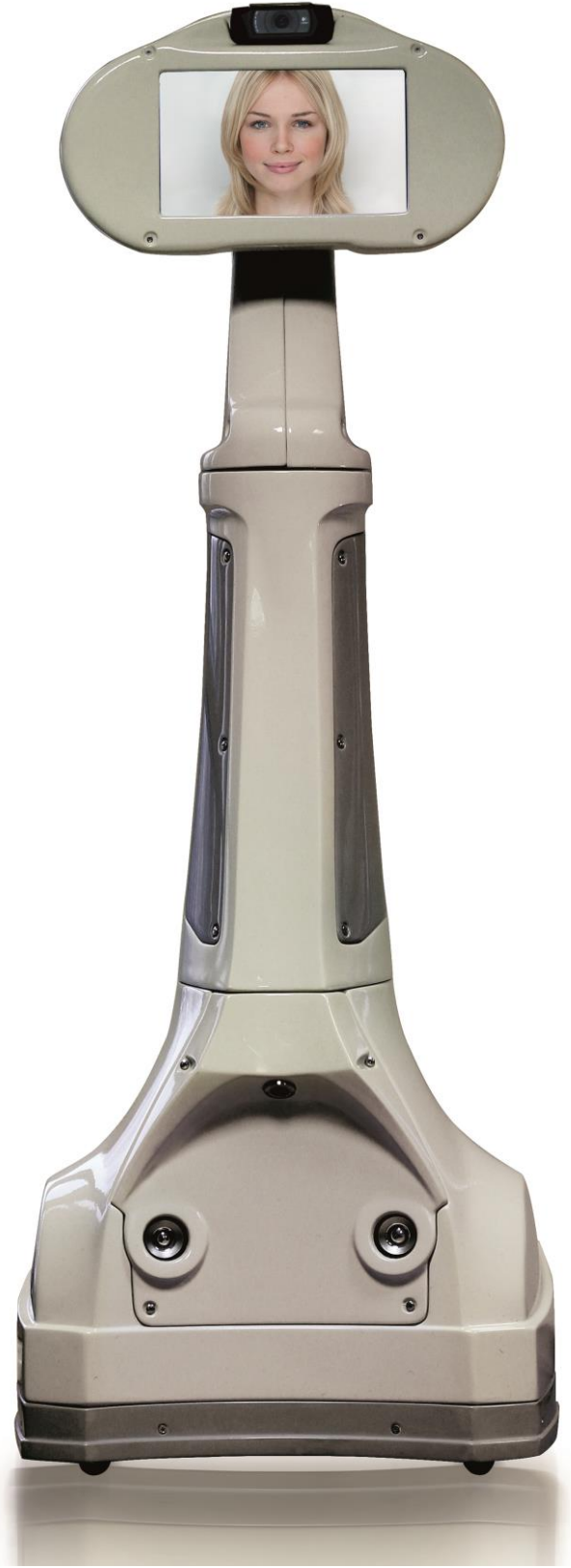

**Figure 1.** Telepresence robot Webot

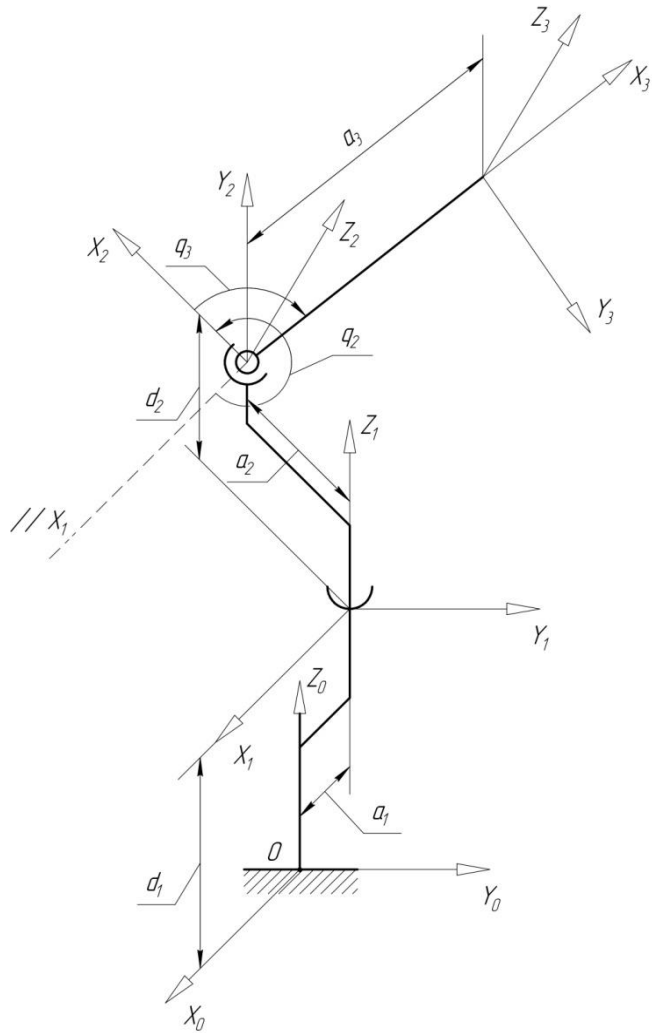

**Figure 2.** Webot kinematic chain

Equations 6-9 demonstrate the calculation of transition matrix T:

$$A_1 = \begin{pmatrix} 1 & 0 & 0 & a_1 \\ 0 & 1 & 0 & 0 \\ 0 & 0 & 1 & d_1 \\ 0 & 0 & 0 & 1 \end{pmatrix} \quad (6)$$

$$A_2(q2) = \begin{pmatrix} \cos q_2 & 0 & \sin(q_2) & a_2 * \cos(q_2) \\ \sin(q_2) & 1 & -\cos(q_2) & a_2 * \sin(q_2) \\ 0 & 0 & 0 & d_2 \\ 0 & 0 & 0 & 1 \end{pmatrix} \quad (7)$$

$$A_3(q3) = \begin{pmatrix} \cos(q_3) & -\sin(q_3) & 0 & a_3 * \cos(q_3) \\ \sin(q_3) & \cos(q_3) & 0 & a_3 * \sin(q_3) \\ 0 & 0 & 1 & 0 \\ 0 & 0 & 0 & 1 \end{pmatrix} \quad (8)$$

$$T(q_2, q_3) = A_1 * A_2(q_2) * A_3(q_3) \quad (9)$$

Equation 1 was written for the case when axis Z and the optical axis of the camera match. Axis Y matches with the optical axis in our case. So we updated equation 1 to arrive at equation 10:

$$\begin{pmatrix} x \\ y \\ \omega \end{pmatrix} = \begin{pmatrix} f_x & 0 & c_x \\ 0 & f_y & c_y \\ 0 & 0 & 1 \end{pmatrix} \cdot \begin{pmatrix} -Z \\ -X \\ Y \end{pmatrix} \quad (10)$$

We implemented an algorithm using C++ language and OpenCV library (Bradski and Kaehler, 2008). We used implementation of reactive navigation system from mrpt library. Our system was able to work in real time on arm7 processor. Also, we implemented user interface for web browser with the help of html5 (Fig. 3).

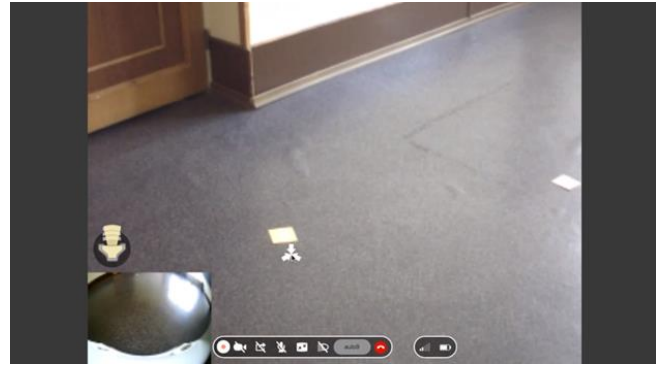

**Figure 3.** Webot user interface

## 4. WEBOT EXPERIMENTS

In the first experiment, Webot, controlled alternately by an operator and by the algorithm we designed, was supposed to travel 2 meters over an open space and stop in a specially marked area (1x0.6 m). Ideally, the robot had to stop in such a way that the center of its body matched the center of the area and the robot retained its spatial angle. During the experiment we measured deltas x, y, f (the distances from the center of the marked area along the corresponding axes and the angle in this coordinate system), and the travel time. These data are shown in Tables 1 and 2.

**Table 1:** Open space, operator control

|    | x, cm | y, cm | f, rad | t, sec |
|----|-------|-------|--------|--------|
| 1  | -6.8  | -9.2  | -0.44  | 10.1   |
| 2  | 5.2   | 12.3  | -0.73  | 9.2    |
| 3  | 15.6  | 13.9  | -0.44  | 9.5    |
| 4  | 3.8   | 11.3  | -0.16  | 9.2    |
| 5  | -7.9  | -7.5  | 0.27   | 7.5    |
| 6  | -6.2  | -6.1  | -0.08  | 9.4    |
| 7  | -1.8  | 3.8   | 0.16   | 9.1    |
| 8  | -3.1  | -11   | 0.11   | 8.9    |
| 9  | 6.7   | 2.1   | -0.59  | 8.4    |
| 10 | -12   | 3.8   | -0.17  | 8.2    |

**Table 2:** Open space, designed algorithm control

|   | x, cm | y, cm | f, rad | t, sec |
|---|-------|-------|--------|--------|
| 1 | 2.8   | 11.8  | -0.38  | 8.7    |
| 2 | 1.5   | 5.4   | -0.06  | 8.9    |
| 3 | -2.4  | -12   | -0.35  | 7.7    |
| 4 | -3.8  | -2.7  | -0.27  | 8.1    |
| 5 | -3.5  | 4.6   | -0.38  | 6.8    |
| 6 | 0.1   | 3.7   | -0.22  | 8.9    |
| 7 | 9.5   | 1.9   | -0.18  | 7.9    |
| 8 | -1.8  | -13.3 | 0.11   | 9.3    |
| 9 | -10 | -9.5 | 0.05 | 8.1 |
| 10 | 0.3 | 2.7 | 0.23 | 8.6 |

In the second experiment, Webot, controlled alternately by an operator and by the designed algorithm, was supposed to travel 2 meters through a doorway and stop in a specially marked area (1x0.6 m). Ideally, the robot had to stop in such a way that the center of its body matched the center of the area. Similar measurements were taken, but the angle value was omitted as irrelevant. The data are demonstrated in Tables 3 and 4.

**Table 3:** Doorway, operator

| | x, cm | y, cm | t, sec |
|---|---|---|---|
| 1 | -8.3 | -18 | 10.9 |
| 2 | 8.9 | -9.2 | 12.3 |
| 3 | 10.1 | 14.4 | 10.4 |
| 4 | 11.4 | 4.6 | 11.3 |
| 5 | 0.6 | 5.2 | 10.3 |
| 6 | 2.2 | 5.5 | 10.7 |
| 7 | -5.6 | 3.9 | 9.4 |
| 8 | -8.1 | 1.5 | 12.2 |
| 9 | 5 | -9.3 | 10.9 |
| 10 | 5.2 | -2.7 | 13.3 |

**Table 4:** Doorway, designed algorithm control

| | x, cm | y, cm | t, sec |
|---|---|---|---|
| 1 | -0.41 | -0.7 | 10.8 |
| 2 | 8.9 | -7.1 | 8.8 |
| 3 | -2.2 | -1.1 | 10.4 |
| 4 | 3.1 | -6.4 | 9.0 |
| 5 | -14.3 | 7.8 | 8.9 |
| 6 | -2.2 | 1.1 | 9.5 |
| 7 | -8.2 | -9.8 | 9.3 |
| 8 | -9.6 | -9.6 | 9.5 |
| 9 | -0.3 | 6.9 | 7.0 |
| 10 | -2.9 | -3.2 | 8.2 |

In the third experiment, Webot, controlled alternately by an operator and by the designed algorithm, was supposed to travel 2 meters, bypassing a 1x1 m block, and stop in a specially marked area (1x0.6 m). Ideally, the robot had to stop in such a way that the center of its body matched the center of the area. We took the same measurements as in experiment 2. The data are presented in Tables 5 and 6.

**Table 5:** Block, operator

| | x, cm | y, cm | t, sec |
|---|---|---|---|
| 1 | -1.5 | -7.3 | 13.8 |
| 2 | 1.4 | -11.2 | 16.9 |
| 3 | -9.3 | -3.4 | 19.1 |
| 4 | -6.9 | 7.4 | 16.7 |
| 5 | -7.2 | 30.9 | 17.6 |
| 6 | -2.5 | 13.1 | 17.8 |
| 7 | -7.9 | -5.1 | 20.3 |
| 8 | 5.1 | -6.6 | 19.7 |
| 9 | 6.0 | -4.2 | 15.5 |
| 10 | -6.7 | -5.9 | 14.5 |

**Table 6:** Block, designed algorithm

| | x, cm | y, cm | t, sec |
|---|---|---|---|
| 1 | -4.0 | 5.4 | 17.7 |
| 2 | -2.7 | -5.7 | 18.9 |
| 3 | -3.6 | -1.7 | 10.9 |
| 4 | -6.4 | -1.2 | 11.5 |
| 5 | 3.0 | 3.7 | 8.9 |
| 6 | 5.3 | -10.1 | 9.2 |
| 7 | -3.7 | 0.0 | 10.5 |
| 8 | -3.2 | 3.6 | 10.1 |
| 9 | -0.2 | 9.8 | 13.3 |
| 10 | 0.9 | 8.1 | 13.0 |

As a result of the experiments, we calculated the percentage that showed the advantage of the control method we designed over a person in solving such tasks. The criteria for the best results were delta values approaching zero (precision of axis and angle positioning) and shorter travel time. The intermediate and final (general) results are shown in Tables 7, 8 and 9 (the final ones are highlighted). The positive figures in the final results of the experiment demonstrate clear advantage of the designed algorithm in controlling the robot.

**Table 7:** Open space

| | x, % | y, % | f, % | t, % |
|---|---|---|---|---|
| 1 | 57 | -32 | 18 | 16 |
| 2 | 53 | 85 | 206 | 3 |
| 3 | 192 | 23 | 28 | 20 |
| 4 | 6 | 106 | -34 | 12 |
| 5 | 63 | 36 | -34 | 8 |
| 6 | 88 | 30 | -43 | 6 |
| 7 | -110 | 23 | -6 | 13 |
| 8 | 19 | -28 | 31 | -4 |
| 9 | -47 | -91 | 166 | 3 |
| 10 | 168 | 14 | -18 | -4 |
| **Results** | 49 | 17 | 31 | 7 |

**Table 8:** Doorway

| | x, % | y, % | t, % |
|---|---|---|---|
| 1 | 121 | 233 | 1 |
| 2 | 0 | 28 | 32 |
| 3 | 121 | 179 | 0 |
| 4 | 127 | -24 | 21 |
| 5 | -209 | -35 | 13 |
| 6 | 0 | 59 | 11 |
| 7 | -40 | -79 | 1 |
| 8 | -23 | -109 | 25 |
| 9 | 72 | 32 | 35 |
| 10 | 35 | -7 | 46 |
| **Results** | 20 | 28 | 18 |

## ACKNOWLEDGMENTS

This work was assisted by the *Wicron* company. It received grant support from the Skolkovo Foundation. The authors applied for the Russian patent 'Hybrid method of telepresence' (application 08071. date 08.12.2015).